\documentclass[twoside,11pt]{article}

\usepackage[utf8]{inputenc}
\usepackage{mathtools}
\usepackage{jmlr2e}

\usepackage{rotating}

\jmlrheading{1}{2000}{1-48}{4/00}{10/00}{meila00a}{Marina Meil\u{a} and Michael I. Jordan}

\ShortHeadings{Hyperplane bounds for neural feature mappings}{Jimeno Yepes}
\firstpageno{1}

\begin{document}

\title{Hyperplane bounds for neural feature mappings}

\author{\name Antonio Jimeno Yepes \email antonio.jimeno@gmail.com \\
      \addr RMIT University\\
      Melbourne, 3000, Victoria, Australia
       }

\editor{Kevin Murphy and Bernhard Sch{\"o}lkopf}

\maketitle

\begin{abstract}

Deep learning methods minimise the empirical risk using loss functions such as the cross entropy loss.
When minimising the empirical risk, the generalisation of the learnt function still depends on the performance on the training data, the 
Vapnik–Chervonenkis(VC)-dimension of the function and the number of training examples.
Neural networks have a large number of parameters, which correlates with their VC-dimension that is typically large but not infinite, and typically a large number of training instances are needed to effectively train them.

In this work, we explore how to optimize feature mappings using neural network with the intention to reduce the effective VC-dimension of the hyperplane found in the space generated by the mapping.
An interpretation of the results of this study is that it is possible to define a loss that controls the VC-dimension of the separating hyperplane.
We evaluate this approach and observe that the performance when using this method improves when the size of the training set is small.

\end{abstract}


\section{Introduction}

Deep learning methods are quite successful in many fields such as image analytics and natural language processing.
Deep learning uses several stacked layers of neural networks, which are optimised using loss functions such as cross entropy with stochastic gradient descent.
An example is presented in eq.\ref{eq:empirical-risk}, where
$\alpha$ represents the possible configurations of the learning machine, $z_{i}$ is a set of examples $i = 1, ..., l$ and $\mathcal{Q}$ is a loss function.
Minimisation of the empirical risk equals minimising eq.~\ref{eq:empirical-risk}.

\begin{equation}
R_{emp}(\alpha) = \frac{1}{l} \sum_{i=1}^l \mathcal{Q}(z_{i},\alpha)
\label{eq:empirical-risk}
\end{equation}

We could consider several ways in which to reduce the risk and several advances have been done in improving stochastic gradient descent~\cite{kingma2014adam}.
Neural networks have a large number of neurons, which implies that they have a large capacity able of modeling a large set of problems and several sets of network weight values could be identified during learning that have minimal empirical risk, which might have different generalisation capabilities.
Their capacity in terms of VC-dimension is large due to the large number of neurons, even though it is finite.

As indicated in equation~\ref{eq:risk-generalization}, the bound on the generalisation performance of a trained model depends on the performance on the training set $R_{emp}(\alpha_{l})$, the VC-dimension $h$ of the classifier and the size of the number of examples used as training data.

\begin{equation}
R(\alpha_{l}) \leq R_{emp}(\alpha_{l}) + \dfrac{B \mathcal{E} (l) }{2} (1+ \sqrt{1 + \dfrac{4 R_{emp} (\alpha_{l}) }{B \mathcal{E} (l)}}) 
\label{eq:risk-generalization}
\end{equation}

\begin{equation}
    \mathcal{E} (l) = 4 \dfrac{h (ln \dfrac{2l}{h}+1)- ln \dfrac{\eta}{4}}{l}
\end{equation}

Assuming that the empirical risk on the training set is the same for several functions, in order to improve the predictive performance, the function with a lower VC-dimension or the one trained with more data should have a lower risk on the performance on an unseen test set.

For controlling the VC-dimension of a function, constraining the effective VC-dimension has been proposed~\cite{vapnik1998statistical}.
Among existing work, regularisation is a way to constrain the search for functions that follow certain properties.
Linear models have relied on Tikhonov regularisation~\citep{tikhonov1977solutions}.
Neural networks are made of a set of non-linearities, thus regularisation such as $L_2$ have been applied~\cite{elsayed2018large}.
There have been several proposals, which include recent work to the models using knowledge~\citep{roychowdhury2021regularizing}.

The structural risk minimisation framework~\cite{vapnik1998statistical} intends to control parameters that minimise the VC-dimension and offers certain guarantees about the performance of the trained model.
A good example of learning algorithm that implements the structural risk minimisation is support vector machines (SVM)~\citep{vapnik2013nature}.
SVM is a large margin classifier that aims separating the classes defining it as a constraint problem, whose solution builds on the Karush–Kuhn–Tucker approach that generalises the Lagrange multipliers.
The vectors from the training set that are on the margin are the named the support vectors.
If we have a training set defined by pairs $\{x_i, y_i\}$ where $x_i$ is a vector in $R^n$ and $y_i = \{1,-1\}$ defines the class of the instance where $i=1,...,l$.
Where the margin is defined by the $1/w \in R^n$ and $b \in R^1$ is the bias term, the constraint optimisation problem for SVMs is defined as:

\begin{equation}
\begin{aligned}
min & & \frac{1}{2}w^2 \\
\\
s.t. & & y_i (w x_i + b) \geq 1
\end{aligned}
\end{equation}

Using Lagrange multipliers $\alpha = \{ \alpha_1,..., \alpha_l \}$ with $\alpha \geq 0$ and $\sum_{i=1}^l \alpha_i y_i = 0$, the separating hyperplane has the formulation:

\begin{equation}
    f(x,\alpha) = \sum_{i=1}^l y_i \alpha_i (x_i*x) + b
\label{eq:svm-hyperplane}
\end{equation}

Eq.~\ref{eq:svm-hyperplane} depends on the inner product of $x_i$ and $x$. The elements of the training set on the margin are named the support vectors.

If the data is not linearly separably in input space, a mapping into a feature space in which the data is linearly separable and an inner product exists would be a solution. For instance, in eq.~\ref{eq:mapping-svm} function $z$ maps the input space into a feature space.
As we can see, what is important is the calculation of the inner product and not the dimensionality of the space.

\begin{equation}
    f(x,\alpha) = \sum_{i=1}^l y_i \alpha_i (z(x_i)*z(x)) + b
\label{eq:mapping-svm}
\end{equation}

From the formulation of SVMs, the inner product between vectors is what is needed to define the solution to the large margin classifier.
Using Mercer's theorem allows using kernels to calculate the inner product in a Hilbert space.
Eq.~\ref{eq:kernel-svm} shows how the SVM formulation would be defined using kernel $K$.

\begin{equation}
    f(x,\alpha) = \sum_{i=1}^l y_i \alpha_i K(x_i,x) + b
\label{eq:kernel-svm}
\end{equation}

Using kernels in this way allows working in high-dimensional feature spaces without having to work in the high-dimensional space explicitly, this is known as the \textit{kernel trick}.
Specialised kernels have been developed to use SVMs in several problem types.


In this paper, we explore using non-linear functions similar to deep neural networks as mapping functions from input space to feature space.
We show examples of the expected performance based on structural risk minimisation principles.
We evaluate the proposed method on several data sets and baseline methods.
When the training data is small, the proposed method largely improves over the baseline methods.
As expected, this improvement is reduced when more training data is provided.

The code used in this experiments is available from the following GitHub repository:\\\url{https://github.com/ajjimeno/nn-hyperplane-bounds}

\section{Methods}

In this section, we define a large margin linear classifier and we introduce the structural risk minimisation principle.
Then, we provide a way to define a large margin classifier using a feature space defined by a set of non-linear functions.
These non-linear functions are the equivalent of a deep neural network.

\subsection{Large margin classifier}



There are several kernels that have been developed over time that turn the input space into a feature space that captures relations among the features, from image analytics (e.g.~\cite{szeliski2010computer,camps2006composite}) to text analytics (e.g. string kernel~\cite{lodhi2002text}).
The kernel trick mentioned above allows working in high dimensional feature spaces without the cost of working directly in the feature space.
This is achieved by using the kernel to calculate the dot product in feature space without having to map the instances into the feature space.
On the other hand, it is difficult to design a kernel that will work well with all sorts of data when compared to the recent success of deep learning.


Deep neural network seem to be effective at approximating many different functions, thus it is interesting to map our input feature space into a feature space in which an optimal hyperplane could be identified.
This means, using a neural network $z$ as the mapping function between the input space and the feature space, so the optimisation problem should consider now $z(x_i) \in R^m$ instead of $x_i$ and now the constraints look like $y_i (w z(x_i) + b) \geq 1$. and $w \in R^m$.

One problem is that current neural networks have a large number of parameters, which are needed to be effective in the current tasks where they are successful.
This implies as well that they have a large capacity or VC-dimension.
In the next sections, we explore how to search for the mapping that has better generalisation guarantees. 







\subsection{Hyperplane bounds}


In this section, we explore properties of the separating hyperplane and what constrains are needed to identify a configuration of the neural network used as mapping function that has better generalisation properties.


If we consider a training set $\{Y,X\}$, as defined above, the following inequality holds true for vector $w_0$ and $\rho_0$,

\begin{equation}
    min_{(y,x) \in \{Y,X\}} \frac{y(w_0 x)}{|w_0|} \geq \rho_0
\end{equation}

which assumes that the training set is separable by a hyperplane with margin $\rho_0$, then the following theorem holds true.


\paragraph{Novikoff theorem}

Given and infinite sequence of training examples $u_i$ with elements satisfying the inequality $|x_i| < D$, there is a hyperplane with coefficients $w_0$ that separates the training examples currently and satisfies conditions as above.
Using an iterative procedure, e.g. stochastic gradient descent, to construct such a hyperplane takes at most

\begin{equation}
    M = [\frac{D^2}{\rho^2_0}]
\end{equation}

As a follow up theorem, we accept that for the algorithm constructing hyperplanes in the regime that separates training data without error, the following bound on error rate is valid

\begin{equation}
    ER(w_l) \leq \frac{E[\frac{D^2_l}{\rho^2_l}]}{l+1}
\end{equation}

where $[\frac{D^2_l}{\rho^2_l}]$ is estimated from training data.

These two theorems provide already a bound on the error of the separating hyperplane, which relies on parameters that can be estimated from the training data.
In the following two sections, we show theorems for the bounds on the VC-dimension of the separating hyperplane and properties of the optimal separating hyperplane that will be used to define the optimisation problem for the neural network mapping function.  

\subsection{Bounds on the VC-dimension for \texorpdfstring{$\Delta$-margin}{TEXT} separating hyperplanes}

In this section, we explore theorems that define bounds on the VC-dimension of the separating hyperplane.

\paragraph{Theorem} Let vectors $x \in X$ belong to a sphere of radius $R$, the set of $\Delta$-margin separating hyperplanes has the following VC-dimension h bounded by the inequality

\begin{equation}
 h \leq min([\frac{R^2}{\Delta^2}], n)+1   
\end{equation}

\paragraph{Corollary} With probability $1-\eta$ the probability that a test example will not be separated correctly by the $\Delta$-margin hyperplane has the bound

\begin{equation}
    P_{error}\leq \frac{m}{l} + \frac{\xi}{2}(1 + \sqrt{1+\frac{4m}{l\xi}})
\end{equation}

where 

\begin{equation}
    \xi = 4 \frac{h(ln \frac{2l}{h} + 1) -ln \frac{\eta}{4}}{l}
\end{equation}

where $m$ is the number of examples not separated correctly by this $\Delta$-margin hyperplane, $h$ is the bound in the VC-dimension, where a good generalisation is dependent on $\Delta$.

So, we have already a bound on the VC-dimension of the separating hyperplane.
In the next section, we present the idea of identifying the optimal hyperplane, which links the hyperplane optimisation problem with structural risk minimisation.

\subsection{Optimal hyperplane properties}

The optimal hyperplane is the one that separates the training examples from classes $y={1,-1}$ with the maximal margin.
It has been previously shown that the optimal hyperplane is unique~\cite{vapnik1998statistical}.
The optimal hyperplane has some interesting properties that are relevant to our work.
One of them is that the generalisation ability of the optimal hyperplane is better than the general bounds obtained for methods that minimise the empirical risk.
Let's define the set $X = {x_1, ..., x_l}$ in space $R^n$, where we have our training examples. Within the elements of $X$ that have the following property:

\begin{equation}
    \inf_{x \in X} | w x + b | = 1
\end{equation}

These elements $x \in X$ are the essential support vectors and are on the margin.
Having defined the essential support vectors, we define the number of essential support vectors $K_l$

\begin{equation}
    K_l = K((x_1, y_1), ... (x_l, y_l))
\end{equation}

And the maximum norm $D_l$ of the essential support vectors.

\begin{equation}
D_l = D((x_1, y_1), ... (x_l, y_l)) = \max_i |x_i|
\end{equation}

Based on the definitions above for the essential support vectors, the following properties have been proved for the optimal hyperplane.
The following inequality $K_l \leq n$ holds true, which implies that the number of essential support vectors is smaller than the dimensionality of the elements of $X$ and $w$.
Let $ER(\alpha_l)$ be defined as the expectation of the probability of error of the optimal hyperplane defined using the training data, then the following inequalities hold for the optimal hyperplane considering the values estimated on the essential support vectors. 


\begin{equation}
    ER(\alpha_l) \leq \frac{EK_{l+1}}{l+1}
\end{equation}


Additionally, considering an optimal hyperplane passing through the origin:

\begin{equation}
    ER(\alpha_l) \leq \frac{E(\frac{D_{l+1}}{\rho_{l+1}})^2}{l+1}
    \label{eq:novikoff-error-expectation}
\end{equation}

Combining the two previous inequalities, we obtain a bound on the expectation of the probability of error, which is bounded by the number of examples in the training data but as well by the number of essential support vectors and the relation between the ball in which the support vectors are and the margin, as shown below

\begin{equation}
    ER(\alpha_l) \leq \frac{E \min(K_l, (\frac{D_{l+1}}{\rho_{l+1}})^2)}{l+1}
\end{equation}

Leave-one-out error has been used as an unbiased estimator to prove the bounds on the optimal hyperplane~\citep{luntz1969estimation}.

\begin{equation}
    E\frac{\mathcal{L}(z_1,...,z_{l+1})}{l+1}=ER(\alpha_l)
\end{equation}

First, the number of errors by leave-one-out does not exceed the number of support vectors~\citep{vapnik2000bounds}.
If a vector $x_i$ is not an essential support vector, then there is an expansion of the vector $\phi_0$ that defines the optimal hyperplane that does not contain the vector $x_i$.
The optimal hyperplane is unique, so removing this vector from the training set does not change it.
Leave-one-out recognizes correctly all the vectors that are not in the set of essential support vectors.
The number $\mathcal{L}(z_1,...,z_{l+1})$ of errors in leave-one-out does not exceed $\mathcal{K}_{l+1}$, which implies


\begin{equation}
    ER(\alpha_l) = \frac{E \mathcal{L}(z_1,...,z_{l+1})}{l+1} \leq \frac{EK_{l+1}}{l+1}
\end{equation}

To prove~eq.~\ref{eq:novikoff-error-expectation}, the number of errors in leave-one-out does not exceed the number of corrections M to find the optimal hyperplane, as defined by Novikoff's theorem presented above. 






\subsection{Mapping into feature space using neural networks}

Up to this point, the formulations rely on an input space defined by the training data instances $x_1, ..., x_l$ in a given space $R^n$.
If a hyperplane separating the data instances into classes $y=\{1,-1\}$ does not exist in input space, the instances are mapped into a feature space $R^m$.

\begin{equation}
    z(x) : R^n \mapsto R^m
\end{equation}

In the case of support vector machines, the~\textit{kernel trick} is used to calculate the dot product using a kernel in feature space without having to do the explicit mapping, which allows working with feature spaces of larger dimensions, even infinite dimensions.
Kernels have been designed to map the input space into separable feature spaces.

Once the kernel is designed, there is a point in the feature space for each training instance.
The properties described for the optimal hyperplane mentioned in the previous sections would still apply but in this case, these properties would be applied to the generated feature space.
Specific formulations are applied to profit from the kernel trick mentioned above.

Developing the best kernel for a specific problem has shown to be a difficult task and multiple kernels have appeared for different tasks.
Recently, neural networks have shown that they can learn a classifier with relatively less effort, despite the increase in computational power required to train these classifiers.
Neural networks in a sense map the input space to another space using a concatenation of linear operators followed by a non-linearity (e.g. sigmoid, RELU, ...), as shown in equation~\ref{eq:neural-network}.

\begin{equation}
z(x) = \sigma f_{k}(\sigma f_{k-1}(... \sigma f_1(x))))
\label{eq:neural-network}
\end{equation}

Each function $f \in {f_1, ..., f_k}$ will be a derivative of $f(x) = Wx+B$, where $W$ and $B$ are matrices of weights and biases.
These are parameters that need to be optimised for each function.
Stochastic gradient descent is typically used to optimize such functions, thus we prepare the optimisation to use stochastic gradient descent.

Revisiting the classification that we intend to optimise and considering the mapping function $z$, we obtain eq.~\ref{eq:neural-network-svm}.
The parameters to optimize are the vector $w$, the bias $b$ and the weights and biases of the neural network $z$.
The dimension of the weight vector $w$ is defined by $z$.

\begin{equation}
y_i(w z(x_i) + b) \geq 1
\label{eq:neural-network-svm}
\end{equation}

In a support vector machine in feature space, a way to compare different kernels is to measure the maximum radius $D_l$ in which the support vectors fit in and multiply it by $\lVert w_{l} \rVert ^2$ as in $[ D_l^2  \lVert W_{l} \rVert^2 ]$.
Considering that $D_l^2 \lVert W_{l} \rVert^2 = \frac{D_l^2}{\rho_l^2}$, the lower is this ration, the lower is the probability of error, which is considered as a mean to control the VC-dimension of the separating hyperplane for the setup defined in this work.





\subsection{Optimisation}

To find the values of the parameters of the system, we use stochastic gradient descent.
Other approaches, such as Lagrange multipliers are not applicable to neural networks.
On the other hand, using stochastic gradient descent guarantees finding a local optima, which hopefully is an approximation with reasonable generalisation performance.

For optimisation, we use AdamW~\citep{loshchilov2017decoupled} with eps=1e-08 and weight decay set to 0.1 (which already implies using an $L_{2}$ regularisation) and betas=(0.9, 0.999).

We have adapted~\citep{zhang2004solving} using large margin loss.
We use the modified Huber loss, which has nicer properties, even though other large margin losses could be explored.

Modified Huber loss~\citep{zhang2004solving}
\begin{equation}
    l(y)=
    \begin{cases}
    -4*h*y   & \text{if }h*y <= -1\\
    (1-h*y)^2  & \text{if } -1 < h*y <= 1\\
    0          & \text{if } h*y > 1
    \end{cases}
\end{equation}

Gradient of the modified Huber loss:

\begin{equation}
    \frac{\partial l}{\partial w_i}=
    \begin{cases}
    -4*h*x_i   & \text{if }h*y <= -1\\
    -2* (1-h*y)*h*x_i  & \text{if } -1 < h*y <= 1\\
    0          & \text{if } h*y > 1
    \end{cases}
\end{equation}

We need to estimate $w$ and $z$ subject to the constraints above.
$z$ will be defined using a neural network and the size of the feature space derived from $z$ will define the size of the vector $w$.
$\lVert w \rVert$ will be minimised and so will be the $\lVert z(x) \rVert$ of the vectors. 
The loss function of the optimisation problem is defined as follows

\begin{equation}
\mathcal{L}(x_1, ..., x_l)_{inv}= \mathcal{L} (x_1, ..., x_l) + \alpha \lVert w \rVert^2 + \beta \sum^l_{i=1} \lVert z(x_i) \rVert^2
\end{equation}

The development above is prepared for binary classification.
In a multi class setting, as many classifiers $y \in \{1, -1\}$ as classes are trained and during prediction, the classifier with the maximum value is returned as prediction as shown in equation \ref{eq:argmax-multi-class}.

\begin{equation}
    \arg \max_{c \in 1..n} \{ f_{1}(x), ..., f_{n}(x) \}
\label{eq:argmax-multi-class}
\end{equation}

The proposed method works in the multi class setting, but it has the advantage that it might be able of deciding when an instance does not belong to any class if all the classifier functions predict the -1 class.


%
%
%
%

\section{Results}

We have evaluated the proposed method using the MNIST and CIFAR 10 data sets.
The different methods have been evaluated using several algorithms.
$\alpha$ and $\beta$ of the loss function have been set to the same value which is the best set up during the experiments.
We evaluate the both losses and the performance of dropout~\citep{srivastava2014dropout} and augmentation based on affine transformations.

\subsection{MNIST}

MNIST is a collection of hand written digits from 0 to 9.
The training set has a total of 60k examples while the test set contains a total of 10k examples.
All images are 28x28 with black background pixels and different white level pixels to define the numbers using just one channel.
Images were normalised using a mean of 0.1307 and standard deviation of 0.3081.

We have used LeNet as base neural network, which has been adapted to be used in our approach.
In order to use LeNet in our approach, we have considered the last layer as the margin weights $w$, which defines each each one of the 10 MNIST classes functions.
The rest of the network has been considered as the function $z(x_i)$.
This means that both the vector $w$ and $z(x_i)$ belong to $R^{84}$.
If the parameters $\alpha$ and $\beta$ are set to zero, we are effectively using LeNet.

Table~\ref{tab:mnist-results-percentage} shows the results of different experimental setups, which includes dropout $p=0.5$, augmentation and a combination of several configurations.
Augmentation was done using random affine transformations with a maximum rotation of 20 degrees, translation maximum of 0.1 on the x and y axis and scale between 0.9 and 1.1, as well brightness and contrast were randomly altered with a maximum change of 0.2.
We have used several partitions of the training set to simulate training the different configurations with data sets of several sizes, which evaluates as well the impact of the number of training examples in addition to the bounded VC-dimension of the separating hyperplane.
Experiments have been run 5 times per configuration and results show the average and the standard deviation.



\begin{sidewaystable}
\begin{tabular}{l|c|c|c|c|c|c|c|c}
\hline
Method&1&5&10&20&40&60&80&100 \\
\hline
ce&91.52$\pm$0.28&97.20$\pm$0.16&98.07$\pm$0.10&98.58$\pm$0.10&99.04$\pm$0.10&99.20$\pm$0.06&99.25$\pm$0.05&99.20$\pm$0.03 \\
ce+do&94.36$\pm$0.22&97.78$\pm$0.09&98.54$\pm$0.07&98.81$\pm$0.07&99.17$\pm$0.03&99.22$\pm$0.03&99.32$\pm$0.03&99.37$\pm$0.04 \\
ce+lm-0.001&94.89$\pm$0.23&97.95$\pm$0.10&98.42$\pm$0.09&98.61$\pm$0.08&99.05$\pm$0.10&99.01$\pm$0.03&99.22$\pm$0.07&99.25$\pm$0.08 \\
ce+lm-0.001+do&95.12$\pm$0.17&97.84$\pm$0.15&98.53$\pm$0.08&98.77$\pm$0.04&98.98$\pm$0.06&99.13$\pm$0.02&99.22$\pm$0.05&99.19$\pm$0.04 \\
\hline
ce+aug&95.82$\pm$0.17&98.37$\pm$0.09&98.94$\pm$0.09&99.23$\pm$0.08&99.36$\pm$0.04&99.40$\pm$0.05&99.52$\pm$0.04&99.50$\pm$0.02 \\
ce+aug+do&95.18$\pm$0.25&97.96$\pm$0.10&98.56$\pm$0.07&98.89$\pm$0.11&99.11$\pm$0.07&99.09$\pm$0.06&99.16$\pm$0.05&99.22$\pm$0.06 \\
ce+aug+lm-0.001&96.82$\pm$0.17&98.68$\pm$0.12&99.08$\pm$0.09&99.24$\pm$0.06&99.37$\pm$0.05&99.41$\pm$0.04&99.47$\pm$0.06&99.53$\pm$0.01 \\
ce+aug+lm-0.001+do&95.57$\pm$0.36&97.91$\pm$0.09&98.55$\pm$0.08&98.78$\pm$0.06&98.95$\pm$0.06&98.99$\pm$0.04&99.05$\pm$0.06&99.16$\pm$0.05 \\
\hline
\hline
mh&91.66$\pm$0.48&97.22$\pm$0.22&98.03$\pm$0.15&98.49$\pm$0.08&98.99$\pm$0.08&99.13$\pm$0.03&99.16$\pm$0.03&99.27$\pm$0.06 \\
mh+do&94.75$\pm$0.32&97.99$\pm$0.10&98.51$\pm$0.04&98.82$\pm$0.05&99.16$\pm$0.05&99.22$\pm$0.08&99.33$\pm$0.04&99.36$\pm$0.08 \\
mh+lm-0.001&94.17$\pm$0.25&97.53$\pm$0.12&98.26$\pm$0.13&98.57$\pm$0.10&98.89$\pm$0.04&98.97$\pm$0.07&99.05$\pm$0.05&99.23$\pm$0.04 \\
mh+lm-0.001+do&95.03$\pm$0.31&97.95$\pm$0.13&98.50$\pm$0.13&98.79$\pm$0.07&99.01$\pm$0.09&99.14$\pm$0.03&99.14$\pm$0.07&99.12$\pm$0.08 \\
\hline
mh+aug&96.20$\pm$0.18&98.45$\pm$0.05&98.96$\pm$0.03&99.22$\pm$0.04&99.40$\pm$0.06&99.44$\pm$0.02&99.53$\pm$0.04&99.51$\pm$0.04 \\
mh+aug+do&95.25$\pm$0.45&98.08$\pm$0.13&98.60$\pm$0.07&98.76$\pm$0.07&99.01$\pm$0.04&99.11$\pm$0.06&99.19$\pm$0.03&99.22$\pm$0.04 \\
mh+aug+lm-0.001&96.53$\pm$0.12&98.62$\pm$0.13&99.05$\pm$0.08&99.19$\pm$0.06&99.41$\pm$0.02&99.46$\pm$0.04&99.46$\pm$0.06&99.50$\pm$0.01 \\
mh+aug+lm-0.001+do&95.14$\pm$0.43&98.03$\pm$0.06&98.52$\pm$0.09&98.78$\pm$0.12&98.90$\pm$0.09&99.00$\pm$0.06&99.09$\pm$0.09&98.99$\pm$0.06 \\
\hline
\end{tabular}
\caption{MNIST results using LeNet using cross-entropy (ce) vs Modified Huber Loss (ml), dropout (do), hyperplane bound loss factor (lm) and augmented (aug)}
\label{tab:mnist-results-percentage}
\end{sidewaystable}

\subsection{CIFAR10}

CIFAR10 is a data set of 32x32 images with 10 classes of objects with 3 channels. There a total of 60k training images and 10k testing images, with an equal split between images.
Each image channel was normalised using a mean of 0.5 and standard deviation of 0.5.

For CIFAR10, we have considered the vgg19~\citep{simonyan14vgg19} network.
The last layer is considered the margin weights $w$ and the output of the rest of the network as the function $z(x_i)$, which belong to $R^{4096}$. 


\begin{sidewaystable}
\begin{tabular}{l|c|c|c|c|c|c|c|c}
\hline
Method&1&5&10&20&40&60&80&100 \\
\hline
ce&28.77$\pm$1.74&43.60$\pm$0.76&51.51$\pm$0.90&60.98$\pm$0.61&70.99$\pm$0.64&76.23$\pm$0.49&78.98$\pm$0.60&81.16$\pm$0.33 \\
ce+do&29.92$\pm$1.22&43.54$\pm$0.94&52.05$\pm$0.75&61.52$\pm$0.81&71.11$\pm$0.68&75.99$\pm$0.68&79.15$\pm$0.56&80.96$\pm$0.37 \\
ce+lm-1e-05&29.73$\pm$1.57&44.01$\pm$0.80&51.54$\pm$0.98&61.11$\pm$0.91&70.92$\pm$0.89&76.34$\pm$0.28&79.68$\pm$0.48&81.23$\pm$0.21 \\
ce+lm-1e-05+do&30.55$\pm$1.46&43.21$\pm$1.36&51.37$\pm$1.28&61.15$\pm$0.73&71.16$\pm$0.24&76.20$\pm$0.55&79.64$\pm$0.58&81.45$\pm$0.56 \\
\hline
ce+aug&32.50$\pm$1.44&49.81$\pm$1.00&61.55$\pm$0.65&70.58$\pm$0.59&79.18$\pm$0.32&83.03$\pm$0.19&85.47$\pm$0.16&87.09$\pm$0.16 \\
ce+aug+do&32.77$\pm$1.53&51.53$\pm$0.92&61.38$\pm$0.59&71.00$\pm$0.48&79.22$\pm$0.32&82.91$\pm$0.29&85.34$\pm$0.27&87.22$\pm$0.19 \\
ce+aug+lm-1e-05&32.53$\pm$1.56&50.03$\pm$1.02&60.85$\pm$0.73&70.81$\pm$0.27&79.38$\pm$0.38&83.21$\pm$0.31&85.25$\pm$0.22&87.12$\pm$0.14 \\
ce+aug+lm-1e-05+do&33.31$\pm$1.27&50.94$\pm$1.93&61.29$\pm$0.63&71.07$\pm$0.64&79.50$\pm$0.39&83.14$\pm$0.32&85.66$\pm$0.29&87.38$\pm$0.14 \\
\hline
\hline
mh&30.26$\pm$1.59&45.11$\pm$1.40&54.14$\pm$0.93&63.39$\pm$0.49&71.93$\pm$0.71&76.83$\pm$0.64&79.49$\pm$0.48&81.58$\pm$0.20 \\
mh+do&30.63$\pm$1.43&46.16$\pm$1.01&54.66$\pm$0.99&63.70$\pm$0.94&71.99$\pm$0.22&76.51$\pm$0.54&79.13$\pm$0.46&81.54$\pm$0.32 \\
mh+lm-1e-05&30.72$\pm$1.35&45.22$\pm$0.96&53.80$\pm$1.19&63.46$\pm$0.80&71.89$\pm$0.59&76.78$\pm$0.52&79.35$\pm$0.56&81.14$\pm$0.37 \\
mh+lm-1e-05+do&31.47$\pm$1.22&45.71$\pm$0.96&53.98$\pm$0.77&62.86$\pm$0.82&71.56$\pm$0.67&76.43$\pm$0.50&79.29$\pm$0.27&81.22$\pm$0.51 \\
\hline
mh+aug&37.25$\pm$0.84&54.38$\pm$0.42&63.54$\pm$0.57&71.64$\pm$0.45&79.66$\pm$0.43&82.98$\pm$0.40&85.25$\pm$0.27&87.10$\pm$0.36 \\
mh+aug+do&36.37$\pm$0.35&54.66$\pm$0.66&63.11$\pm$0.69&72.06$\pm$0.53&79.46$\pm$0.29&83.06$\pm$0.40&85.32$\pm$0.18&87.03$\pm$0.21 \\
mh+aug+lm-1e-05&36.91$\pm$0.98&54.54$\pm$0.76&63.53$\pm$1.00&71.35$\pm$0.51&79.49$\pm$0.41&82.90$\pm$0.37&85.36$\pm$0.21&86.94$\pm$0.23 \\
mh+aug+lm-1e-05+do&36.30$\pm$1.34&54.61$\pm$0.70&63.34$\pm$0.60&72.08$\pm$0.37&79.44$\pm$0.40&83.14$\pm$0.21&85.45$\pm$0.34&87.09$\pm$0.16 \\
\hline
\end{tabular}
\caption{CIFAR10 results using vgg19 cross-entropy (ce) vs Modified Huber Loss (ml), dropout (do), hyperplane bound loss factor (lm) and augmented (aug).}
\label{tab:cifar-results-percentage}
\end{sidewaystable}

\section{Discussion}

We have evaluated the proposed approach using two well-known image classification data sets and compared againt several baselines.
The results are compared against several baselines, which includes dropout and augmentation using affine transformations.
In addition to the modified Huber loss, we compare as well the performance of cross entropy loss, which is typically used in deep learning.

We have mentioned at the beginning that there are two factors that are relevant for the risk of generalisation of machine learning algorithms.
One is the VC-dimension, which we have tried to influence in this work.
The second one is additional training data.
Providing additional training data has been simulated in two ways.
The first one is by selecting portions of the training data available from the training sets.
The second one has been using affine transformations, which uses the examples in the portion of the data set used as training data.

We observe that the proposed method shows a strong performance improvement when less training data is being used.
We observe as well that by providing additional training data effectively affects performance on the test set positively, which can be combined with the proposed method.
Dropout performed well when no modification or data augmentation were used.

When considering the values of $D_l^2$ and $\lVert w_l \rVert^2$ in the loss function, the changes in size on the training set seems to have limited impact. The values for the margin are very similar, while the radius $D_l^2$ for the support vectors are around 1.
Considering these variables in the loss function, their values become more stable between training runs.
When not using these terms as part of the loss function, the values tend to change significantly between experiments.

\section{Related work}

Deep learning has achieved tremendous success in many practical tasks but
the neural networks used in deep learning have a high number of parameters, which implies a high VC-dimension.
These networks are optimised to reduce the empirical risk and having large data sets helps reducing the generalisation risk linked to this.

Neural networks are trained using stochastic gradient descent, which guarantees reaching an optimum even if it is a local one.
There is recent work in which the neural networks is studied within the kernel theory, which has shown interesting understanding about the global optimisation of neural networks~\citep{du2019gradient} and the convergence of wide neural networks through
Neural Tangent Kernels~\citep{jacot2018neural}.
This work complements what we have studied in this work and provides as well directions for further research.
On the other hand, this previous research could be considered to define directions for better defining approximation functions that would use well know properties of kernels and the adaptability of neural networks. 

Compared as well to what we propose in this work from the point of view of regularisation, there have been several means of regularisation of neural networks that include $L^2$ regularisation and several variants where different layers are regularised independently.
Transfer learning is a way to pre-train the neural networks using similar data to the training data, which has been quite successful in recent proposed systems to improve the capability of existing systems.
In addition, knowledge has been proposed as means of regularisation~\citep{borghesi2020improving,roychowdhury2021regularizing}, which would reuse existing resources and reduce the need for training data.

In this work, we have used hyperplane bounds to study how this contribute to find a better hyperplane using neural networks as feature space mapper.
There are several recent directions that would provide directions to improve the definition of functions that have better guarantees in terms of optimisation.

\section{Conclusions and future work}

Improve over the baseline, which is most significant when a small portion of the training data is used.
When more training data is made available, the improvement is reduced, which is an expected behaviour as shown by the bounds of the VC-dimension for $\Delta$-margin separating hyperplanes and formulation of the empirical risk.

We have considered well known networks for the experiments, based on convolution neural networks, it might be interesting to explore additional network configurations to understand the impact of the network architecture with the proposed approach.
There has been as well recent work to better understand the optimisation and behaviour of neural networks that we would like to explore as future work.

\section{Acknowledgements}

This research was undertaken using the LIEF HPC-GPGPU Facility hosted at the University of Melbourne. This Facility was established with the assistance of LIEF Grant LE170100200.

\bibliography{bibliography}

\end{document}